\documentclass[sigconf,protrusion=true,expansion=true,nonacm]{acmart}

\usepackage{amsmath,amsfonts}

\usepackage{amssymb}
\usepackage{algorithmic}
\usepackage{geometry}
\usepackage{graphicx}
\usepackage{algorithm}
\usepackage{textcomp}
\usepackage{xcolor}
\usepackage{array}
\usepackage{multirow}
\usepackage{makecell}
\usepackage{ragged2e}
\usepackage{balance}
\usepackage{verbatim}
\usepackage{fancyhdr}
\setcopyright{none}

\AtBeginDocument{%
  }

\copyrightyear{2024}
\acmYear{2024}
\acmDOI{10.1145/3670474.3685965}
\acmISBN{979-8-4007-0699-8/24/09}

\begin{document}

\title{Cell Library Characterization for Composite Current Source Models Based on  Gaussian Process Regression and Active Learning}
% \title{Cell Library Characterization for Composite Current Source Models Based on GPR and Active Learning}
% \author{Tao Bai, Zeyuan Deng and Peng Cao*}
% \affiliation{
% \institution{National ASIC System Engineering Technology Research Center, Southeast University, Nanjing, China  \\
% National Center of Technology Innovation for EDA, Nanjing, China}
%   \city{}
%   \country{}}

\author{Tao Bai$^{1,2*}$, Junzhuo Zhou$^{3*}$, Zeyuan Deng$^{1,2}$, Ting-Jung Lin$^{4}$, Wei Xing$^{5}$, Peng Cao$^{1,2\dagger}$, Lei He$^{3}$
}

\affiliation{
  \institution{$^1$ National ASIC System Engineering Technology Research Center, Southeast University, Nanjing, China}
  \country{}  % 添加这行
}

\affiliation{
  \institution{$^2$ National Center of Technology Innovation for EDA, Nanjing, China}
  \country{}  % 添加这行
}

\affiliation{
  \institution{$^3$ University of California, Los Angeles, United States}
  \country{}  % 添加这行
}

\affiliation{
  \institution{$^4$ Ningbo Institute of Digital Twin, Eastern Institute of Technology, Ningbo, China}
  \country{}  % 添加这行
}

\affiliation{
  \institution{$^5$ The University of Sheffield, Sheffield, United Kingdom}
  \country{}  % 添加这行
}

\email{caopeng@seu.edu.cn, lhe@ee.ucla.edu}
\renewcommand{\shortauthors}{Tao et al.}

\begin{abstract}
%With process technology scaling, accurate gate-level timing analysis becomes increasingly challenging. Highly resistive on-chip interconnects significantly impact timing, as signals no longer resemble smooth saturated ramps, and gate-interconnect interdependencies grow stronger. Furthermore, with advanced process nodes, the number of process corners required for timing sign-off has dramatically increased, leading to substantial memory and time overheads for characterizing standard cell libraries for each scenario. 

The composite current source (CCS) model has been adopted as an advanced timing model that represents the current behavior of cells for improved accuracy and better capability than traditional non-linear delay models (NLDM) to model complex dynamic effects and interactions under advanced process nodes.
However, the high accuracy requirement, large amount of data and extensive simulation cost pose severe challenges to CCS characterization.
To address these challenges, we introduce a novel Gaussian Process Regression(GPR) model with active learning(AL) to establish the characterization framework efficiently and accurately. 
Our approach significantly outperforms conventional commercial tools as well as learning based approaches by achieving an average absolute error of 2.05 ps and a relative error of 2.27\% for current waveform of 57 cells under 9 process, voltage, temperature (PVT) corners with TSMC 22nm process.
Additionally, our model drastically reduces the runtime to 27\% and the storage by up to 19.5$\times$ compared with that required by commercial tools. 
%This innovative approach maintains superior accuracy and accelerates the characterization process, proving to be highly effective in the face of very deep submicron process challenges.
\end{abstract}


\maketitle

% 设置页眉页脚
% Define the headers for even and odd pages
% \fancypagestyle{main}{
%     \fancyhf{}
%     \fancyhead[LO]{\small Cell Library Characterization for Composite Current Source Models Based\\ on Gaussian Process Regression and Active Learning}
%     \fancyhead[LE]{\small MLCAD ’24, September 9-11, 2024, Salt Lake City, UT, USA}
%     \fancyhead[RO]{\small MLCAD ’24, September 9-11, 2024, Salt Lake City, UT, USA}
%     \fancyhead[RE]{\small \shortauthors}
%     % \fancyfoot[C]{\thepage}
%     \renewcommand{\headrulewidth}{0pt} % remove the header line
% }

% Apply the header styles
% \thispagestyle{firstpage} % first page header
% \pagestyle{main} % main pages header

\let\thefootnote\relax
% \footnotetext{This work was supported in part by the National Natural Science Foundation of China under Grant 62174031 and in part by the Fundamental Research Funds for the Central Universities.}
\footnotetext{$^{*}$Both authors contributed equally to this research.}
\footnotetext{$^{\dagger}$Peng Cao is the corresponding author.}
\section{Introduction}
% 一：
% 1.单元特征化的作用
% 2.工艺节点缩小后，遇到的问题
% 3.CCS能克服上述问题，模型概念，但仍存在问题

Standard cell libraries are essential in integrated circuit design for performance, power, and area (PPA) optimization, which are characterized to enable accurate timing and power analysis during logical synthesis and physical implementation \cite{4786726}.
Different from traditional non-linear delay models (NLDM) using simple delay values to represent the cell behavior, the composite current source (CCS) model is an advanced timing model that represents the current behavior of cells for improved accuracy and better capability to model complex dynamic effects and interactions under advanced process nodes \cite{goyal2005current}.
However, suffering from considerably increased complexity with advanced nodes, the more time-consuming and resource-intensive computational effort is induced as well as dramatically increased storage requirements \cite{9382263,9665573}, which poses a significant challenge for the characterization and timing analysis of CCS model.

%Standard cell libraries are vital in integrated circuit design, significantly reducing development time while enhancing design robustness and flexibility. These libraries cater to diverse design requirements such as area, timing, power, and testability, suitable for both high-speed and low-power circuits. High-quality libraries are essential for effective and optimized RTL implementations.
%The widely available Non-linear delay and power models (NLDM/NLPMs) have long been the method used to abstract delay and output transition time characteristics of cells for usage in timing flows. At process geometries of 90nm and below, many new effects cannot be accurately modeled using this approach. Several modeling challenges arise in the context of digital circuit design, including issues such as High Impedance Interconnect, the Miller effect, Dynamic IR-drop, among other complexities\cite{el2011ccs,9394564}.
%To make matters worse, some of these effects are inter-dependent between timing, noise and power. For example timing and slew rates affect power which impacts IR-drop which in turn changes timing. Also signal integrity can impact power which in-turn impacts IR-drop and timing. With the advent of smaller nanometer leading to the issues discussed above, the composite current source (CCS) approach of modeling cell behavior has been developed to address the new effects of very deep submicron processes. 

% As illustrated in Fig. \ref{fig:CCS}, the CCS model is composed of three main components, 
The CCS model is composed of three main components, including the receiver model, the driver model, and the reduced-order network model \cite{garyfallou2022leveraging}.
The driver model represents the behavior of the output driver of a cell, capturing how it generates currents over time in response to input signals, while the receiver model characterizes how the input of a cell responds to incoming signals, focusing on the input pin's electrical characteristics.
The reduced-order network model simplifies the complex interconnect network between drivers and receivers, making the CCS model more manageable without losing significant accuracy.
Among them, the driver model's accuracy directly impacts the overall timing and noise analysis of the cell, which involves numerous SPICE simulations to generate the current vs. time data for different process, voltage, and temperature (PVT) corners.
Inaccurate driver characterization can lead to incorrect predictions of signal behavior, affecting the reliability of the design.

%the CCS timing modeling is composed of a driver model and a receiver model. The driver model is a time- and voltage-dependent current source. Because this current source’s drive resistance is essentially infinite, the model provides high accuracy even when the drive resistance is much lower than the interconnect impedance. The CCS receiver model consists of two capacitances C1 and C2 \cite{garyfallou2022leveraging}, allowing dynamic adjustment of capacitance during the transition. The capacitance values can also be dependent on input slew, output load and state of the cell.

% 二：CCS delay slew如何计算(5.29)
%Characterization of standard cells based on CCS model, the output current waveform of a cell depicts the progression of the output voltage across the full range up to VDD. The timing information of the cell can be extracted by converting the output current waveform into the output voltage waveform using the trapezoidal integration method.

% 三：现在在K库方面存在什么问题
The increasing complexity of semiconductor technology and the number of required PVT corners have introduced severe challenges in CCS characterization due to the high accuracy requirement, large volume of data, and extensive simulation overhead \cite{siemens}.
First, capturing detailed current waveforms requires high-precision measurements and simulations to accurately reflect the dynamic behavior of the driver during switching. 
Any inaccuracies in these waveforms can lead to incorrect timing and noise predictions \cite{ccs_timing}. 
%[此处找1篇ccs K库论文，老的，方法上普通的即可，没有的话找篇介绍CCS的也行]
Second, since the characterization must be performed across a wide range of input transition times, output loads, and PVT corners, the amount of data is exponentially increased for current waveforms with high temporal resolution for each cell and each condition.
Prior works have been presented to reduce the storage cost by waveform compression, but they focus on the current waveform of single cell only and neglect the correlation among different cell driver strengths and different PVT corners \cite{9665573,5090841}.
Third, the sophisticated characterization scenarios also result in time-consuming simulation costs to capture fine-grained current waveforms, potentially slowing down the design cycle.
In spite of considerable researches focused on accelerating the characterization of statistical and aging-aware timing libraries \cite{naswali2021fast,8965191,9394564,ebrahimipour2020aadam}, few of them were devoted into the area of CCS characterization.

To tackle the aforementioned challenges, an accurate and efficient standard cell library characterization framework is proposed in this work for CCS model based on Gaussian Process Regression (GPR) and active learning (AL), which significantly reduces the data storage requirement and simulation cost with considerable accuracy for multiple PVT corners and cells with varying driving strengths.
The main contributions of this work are summarized as follows.
%we proposed a rapid and accurate Gaussian Process Regression model enhanced with Active Learning, capable of characterizing standard cells across multiple PVT corners and varying drive strengths.
%The contributions of this work are listed as follows,
% 几点长度差不多
% 1.GP
% 2.AL
% 3.跨PVT 

\begin{itemize}
    \item To the best of our knowledge, this is the first work to propose a CCS model characterization framework for standard cell library with GPR and AL, which formulates the detailed modeling of current waveform as the uncertainty quantification (UQ) problem and predicts the individual current and time value with high precision and efficiency.
    \item Considering the uncertainty inconsistency issue, multiple GPR models are employed to improve the prediction precision of the corresponding current points, where the correlation among PVT corners and driving strengths are captured as features to enhance generalization performance and reduce simulation cost.  
    \item The strategy for selecting training data efficiently is implemented by AL, where the samples with the lower contribution to accuracy are selected by UQ and used to retrain GPR models so as to minimize the amount of required training data and accelerate model convergence.

\end{itemize}

The rest of the paper is organized as follows. Section \ref{Pre} introduces the related works of standard cell characterization and preliminaries about GPR and AL, and Section \ref{Framework} introduces the proposed algorithm, and Section \ref{result} discusses our experimental setup and results, followed by conclusion in Section \ref{Conclusion}.

\section{RELATED WORKS AND PRELIMINARIES}
% 第二页
\label{Pre}
\subsection{Related Works}
% 4.加速K库的国内外研究现状
%     a.基于数学方法拟合
%     b.机器学习(GNN,Ridge)
In order to improve accuracy, efficiency, and scalability in the characterization process of standard cell library, plenty of studies have been dedicated to developing modeling methodologies for modern process nodes, 
leveraging machine learning methods, especially supervised learning, to accelerate the establishment of standard cell libraries has become a prominent focus of research. 

The deep neural network (DNN) model is applied to perform standard cell characterization for nominal timing \cite{8965191} and statistical timing \cite{naswali2021fast}, respectively, which use data from sparse characterization to generate delay models at required sign-off corners but ignore the relation with transistor-level cell topology.
A characterization approach using ridge regression for aging-aware cell library is detailed in \cite{9394564}, which enables designers to assess their circuits under precisely selected degradation scenarios. 
%However, this model neglects the timing correlations across different PVT corners and drive strengths. 
Similarly, the work of \cite{8988626} introduced an innovative logical circuit simulation method implemented using neural networks based on the current source model. 
The research in \cite{chen2023unleashing} employs a multi-layer perceptron (MLP) model to create power supply noise (PSN) waveform for each standard cell.
Furthermore, \cite{zhou2025lvfgen} utilizes GPR with AL to estimate statistical timing distributions in OCV characterization.

Although various ML algorithms have been utilized, the method for selecting training data of CCS characterization has not been thoroughly studied. This oversight has a significant impact on prediction accuracy and the simulation cost required to generate the data.
Moreover, the uncertainty of predicted results by ML algorithms may degrade the accuracy of cell library, which is vital for circuit design.

%considers the impact of noise on delay, thus eliminating the need for Look Up Table and enhancing the accuracy of delay calculations. However, the models developed by Abrishami, Zhuo, and the ridge regression approach all share a common limitation: Each of these models necessitates the construction of individual models for each PVT and drive strengths, and fails to account for the interdependencies between different PVT conditions and drive strengths. This lack of consideration for the inherent correlations restricts their practical efficacy and adaptability in varying operational environments.

%In the current VLSI design landscape, supporting multiple PVT corners and drive strengths simultaneously is considered a significant breakthrough. This advancement allows researchers to more comprehensively capture and reflect circuit performance characteristics under a broader range of operational conditions. By accommodating multiple PVT scenarios, the models can better adapt to the uncertainties and variations in the actual working environment. 

%Overall, these studies provide a variety of effective approaches to accelerate the establishment of standard cell libraries based on the CCS model, and they also highlight the importance of more efficient modeling methods for improving the efficiency of standard cell library design as process nodes continue to evolve.

\subsection{Gaussian Process Regression}
% (检查每个符号是否有解释）
From a functional space perspective, a Gaussian process (GP) is defined to describe the distribution of functions, enabling direct Bayesian inference within the function space. A GP is a collection where any finite set of random variables possesses a joint Gaussian distribution. The characteristics of a GP are entirely determined by its mean function, $\mu(x)$, and covariance function, $k(x, x^{\prime})$, as shown in Eq. \ref{m(x)} and \ref{k(x,x')} \cite{murphy2012machine}.
\begin{equation}
\label{m(x)}
\begin{aligned}
& \mu(\boldsymbol{x})=\boldsymbol{E}[f(\boldsymbol{x})]
\end{aligned}
\end{equation}
\begin{equation}
\label{k(x,x')}
\begin{aligned}
& k\left(\boldsymbol{x}, \boldsymbol{x}^{\prime}\right)=\boldsymbol{E}\left[(f(\boldsymbol{x})-\mu(\boldsymbol{x}))\left(f\left(\boldsymbol{x}^{\prime}\right)-\mu\left(\boldsymbol{x}^{\prime}\right)\right)\right]
\end{aligned}
\end{equation}
where \( \mathbf{x}, \mathbf{x}' \in \mathbb{R}^d \) are arbitrary random variables. The kernel function \(k(\mathbf{x}, \mathbf{x}')\) encapsulates the relationship between the function values at points \( \mathbf{x} \) and \( \mathbf{x}' \). Therefore, the GP can be defined as \( f(\mathbf{x}) \sim \text{GP}(\mu(\mathbf{x}), k(\mathbf{x}, \mathbf{x}')) \).

GPR model provides an estimate of uncertainty in predictions by quantifying the reliability of the predictions through confidence intervals. 
In this work, we utilize multiple GPR models to obtain the uncertainty associated with each output.

\begin{figure*}[ht!]
    \centering
    \includegraphics[width=0.95\textwidth]{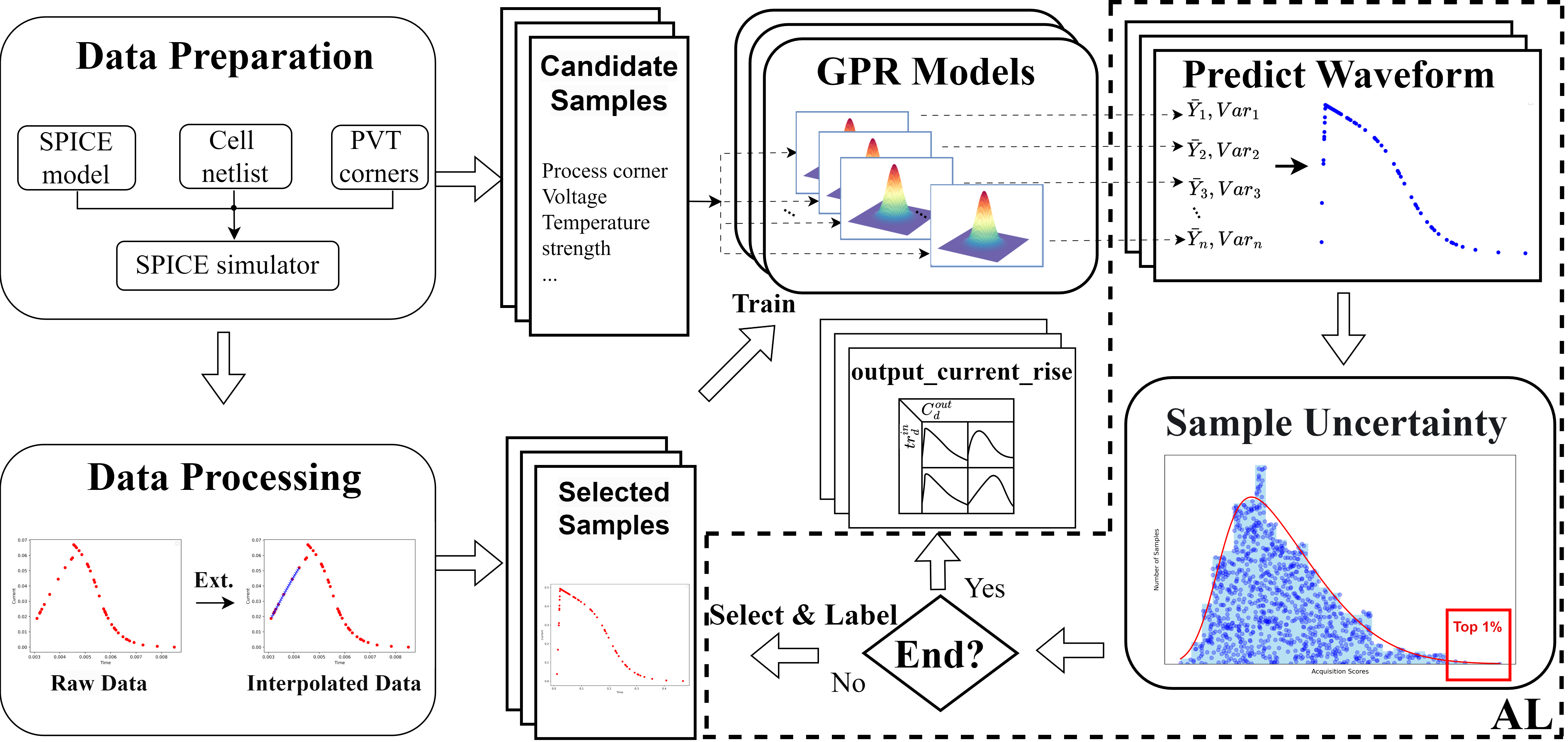} % Use the correct image file name
    \caption{Overview of Proposed Standard Cell Characterization Framework for CCS Model Based on GPR and AL}
    \label{fig:framework}
\end{figure*}

\subsection{Active Learning}
% AL广泛应用
% AL strategies are used in supervised learning to let the training algorithm "ask questions", i.e. choose the feature vectors to query for the corresponding target value during the training phase, usually based on the model learned so far. The main aim of AL is to possibly reduce the number of training samples required to train the model, or in other words, to get a model of the same prediction quality with a smaller dataset. This is particularly useful when knowing the target value associated with a given combination of features, which is an expensive operation. For example, it may involve asking a human to "label" samples manually, running a costly and time-consuming laboratory experiment, or performing a complex computer simulation.

Active Learning (AL) strategies in supervised learning allow the training algorithm to "ask questions" by selecting feature vectors to query for their target values during training, usually based on the current model. The primary goal of AL is to reduce the number of training samples needed to achieve a model with high predictive quality, which is especially valuable when obtaining target values is costly or time-consuming, such as requiring human labeling, expensive experiments, or complex simulations.
AL methods are usually categorized in query synthesis (or population-based) methods, in which the feature vector to query can be chosen arbitrarily, and pool-based sampling methods, in which the vector can only be chosen within a given finite set (or "pool") of unlabeled values, and selective-sampling methods, in which vectors are proposed in a streaming flow and the AL algorithm can only decide online whether to ask for the corresponding target or not.

Several approaches to AL are available in the literature. Most of the literature focuses on classification problems \cite{Aggarwal2014}, although AL has been investigated also for regression \cite{8475012}. In this work, the uncertainty of each output is obtained through the surrogate mode model. From this, the uncertainty of the sample is derived, allowing for the selection of samples that require labeling.
 
% Active learning is a framework in machine learning where the algorithm can query a user or an oracle to label new data points with their true labels, referred to as optimal experimental design. This approach allows the algorithm to proactively select which examples to label from a pool of unlabeled instances. The principle is that by allowing the machine learning algorithm to select the data from which it learns, it could potentially achieve higher accuracy with fewer training labels. This process is often seen as a powerful instance of the Human-in-the-Loop paradigm and is considered a semi-supervised learning method. It strikes a balance between unsupervised learning (using no labeled examples) and supervised learning (using all labeled examples). Active learning can greatly enhance model efficiency by iteratively increasing the labeled dataset size, achieving near fully-supervised performance at a fraction of the cost or training time.

% Active learning is categorized into three primary methods: Membership Query Synthesis, Stream-Based Selective Sampling, and Pool-Based Sampling, each differing in how data is selected for labeling. A crucial aspect of active learning is determining which sample data needs to be labeled, based on the uncertainty of an unlabeled sample. In this work, the uncertainty of each output is obtained through the Gaussian Process Regression model. From this, the uncertainty of the sample is derived, allowing for the selection of samples that require labeling.

\section{CCS Timing Model Characterization Framework}
% 3,4页
\label{Framework}
\subsection{Problem Formulation}
% liberate换掉(已换掉
% 1.问题定义
In this work, the accurate CCS model characterization for each specific function of cells, e.g. NAND cell, is formulated as a UQ problem, whose current waveform is represented under $N$ different conditions as $\mathcal{X} = {x_i | i = 1, 2, ..., N}$.
By exploiting the correlation among multiple PVT corners and different driving strengths, the $N$ conditions for a cell include the combinations for timing arc, input transition and output load.
With the predicted and real current waveform, the corresponding voltage waveforms could be represented as \( V^{(i)} \) and \( V^{(i)*} \), whose difference is used to evaluate the uncertainty of the characterization framework in terms of root mean squared error (RMSE) as shown in follows.
\begin{equation}
L^{(i)} = \operatorname{RMSE}(V^{(i)} - V^{(i)*})
\end{equation}

\iffalse

In order to perform an accurate and efficient CCS timing characterization for standard cell library under multiple corners and drive strength cell,
% a GPR based proposed with active learning in this work. 
%To minimize simulation efforts, we formulate CCS timing cell characterization as an Uncertainty quantification(UQ) optimization problem. 
The samples to be predicted are $\mathcal{X} = {x_i | i = 1, 2, ..., N}$, in the \(i\)-th sample, the aim is to minimize the distance between the estimated voltage waveform and the real voltage waveform, effectively reducing the uncertainty associated with the model's predictions. That is,
\begin{equation}
L^{(i)} = \operatorname{RMSE}(V^{(i)} - V^{(i)*})
\end{equation}
Where \( L^{(i)} \) is the objective function of the i-th sample, representing the distance between the predicted voltage waveform \( V^{(i)} \) and the simulated voltage waveform \( V^{(i)*} \), we use the RMSE to quantify the difference between the two voltage waveforms. 
\fi

%To align the time points between the predicted and simulated waveforms, we interpolate sufficient identical time points within the duration of both waveforms. The corresponding current values are then interpolated, and from these, the corresponding voltage values are calculated. The RMSE between the predicted and simulated voltage waveforms is computed to serve as \( L^{(i)} \). 

Therefore, the objective to maximize the accuracy of voltage waveform for CCS model characterization is formulated as the minimization of \( L^{(i)} \).
%, ensuring that the model's predicted waveform achieves the highest accuracy.

\begin{equation}
\text{Objective: } \min(L^{(i)})
\label{obj}
\end{equation}

\subsection{Overview}
% \flushbottom
% 2.粗略讲，提到图里的大标题，详细的在B里说，写到跟图高差不多（5.29）
%In this work, a GPR) model with AL is proposed to characterize standard cells based on the CCS Model. 
Fig. \ref{fig:framework} demonstrates the overview of the proposed standard cell characterization framework for CCS model based on GPR and AL.
As illustrated in Fig. \ref{fig:framework}, the SPICE model and cell netlist, as well as the information of PVT corners are prepared for SPICE simulation to generate current waveforms as training data.
Considering the different number of current points for specific functions of cells with different driving strengths, linear interpolation operation is performed to extend the original current points to a fixed identical number.
For the task of characterization framework establishment for CCS model, the GPR models are trained and AL with processed training data for selected samples and predict the current waveform for candidate samples with the extracted features according 
\vspace{-0.5mm}
to PVT corner and cell netlist.
During training, based on the predicted waveform for candidate samples, those with the worst uncertainty, e.g. top 1\%, are added into the selected samples to retrain the GPR models repeatedly until the termination criteria are met.
Finally, the GPR models are stored as the characterization framework for CCS model to represent the current waveform of each specific function of cells.

\subsection{Design of an Algorithm Using GPR with AL}
\label{Algorithm}
In this section, we detail a comprehensive algorithmic framework designed for the advanced characterization of standard cells. This framework employs GPR models with AL, structured into a series of systematic steps to ensure robust and precise outcomes. Starting with the initialization and training phases, the process flows through a cycle of prediction, evaluation, and iterative refinement, crucial for handling the dynamic complexities of semiconductor data.

\subsubsection{Initialization.}
\label{init}
In this study, we generate standard cell libraries under various Process, Voltage, and Temperature (PVT) corners and different drive strengths. To maintain consistency across the data set, we extend all data to the same length through linear interpolation without compromising the accuracy of the timing information. We then generate a large set of candidate samples, denoted as $\mathcal{X}_c = {x_i | i = 1, 2, ..., N}$. From this set, we select a subset $\mathcal{X} = {x_j | j = 1, 2, ..., M}$ with $M < N$, where the associated simulated waveform data, $\mathcal{Y} = {y_j | j = 1, 2, ..., M}$, originates from HSPICE simulations. Finally, we normalize the selected samples to facilitate further analysis and modeling.

\subsubsection{Gaussian Process Model Training.}
\label{train}
Using the selected samples $\mathcal{X}$ and their corresponding waveform data $\mathcal{Y}$, we train GPR models to effectively predict and analyze the behavior of standard cells under various conditions.

\subsubsection{Waveform Prediction by Gaussian Process Model.}

We use the trained Gaussian process model to predict the current waveform, leveraging the sophisticated statistical techniques embedded in the model to generate precise waveform outputs from the selected sample data.

\subsubsection{Uncertainty Evaluation.}

The uncertainty of all candidate samples is evaluated using an acquisition function $A^{(i)}$, $i<N$. This acquisition function, which is detailed in Section \ref{Acquisition}, helps quantify the uncertainty associated with each sample's prediction, facilitating more informed decision-making in subsequent steps.

\subsubsection{Termination Check.}

The algorithm performs a termination check to decide whether to continue or stop. If the specified stopping criterion is met, either because the iteration count exceeds the maximum allowed iterations or the highest acquisition score among the candidate samples falls below a predefined threshold, the algorithm is terminated. If neither condition is met, the process continues with the selection and labeling of new samples.

\subsubsection{Selected Samples Update.}
In this step, we select the top 1\% of candidate samples with the highest uncertainty, label them, and add them to the pool of selected samples. These updated samples are then normalized to ensure consistency in data format and scale. Following normalization, the surrogate model is retrained, referring back to step \ref{train}, to incorporate the newly labeled data, thereby enhancing the model's accuracy and reliability.

This algorithmic workflow harnesses the power of GPR combined with AL to provide a state-of-the-art approach for standard cell characterization. The methodology not only achieves high accuracy in waveform prediction but also optimizes the computational resources and time required for simulations.

\subsection{Acquisition by Variational Analysis}
\label{Acquisition}

% 为了分别考虑时间，电流预测的准确性，将A(i)分为两部分

To evaluate the uncertainty of all candidate samples, we compute the uncertainty of each sample by using the acquisition function (Eq. \ref{acquisition function}) to evaluate the uncertainty associated with each prediction made by the GPR models. By selecting the top 1\% of samples with the highest acquisition scores for retraining and iterating through this process several times, the model achieves significantly improved accuracy.

We define the uncertainty in the distance between the converted voltage waveform from the predicted current waveform of the \(i\)-th sample and the simulated voltage waveform as the Acquisition Function. Eq. \ref{acquisition function} quantifies how uncertainty in the GP model’s predictions at different time or current points contributes to the overall uncertainty in the \(i\)-th sample. By multiplying each predicted value's uncertainty with the corresponding gradient and summing these products, we can derive the sample's overall uncertainty. This approach is used to iteratively assess and refine the model’s performance.

% \begin{equation}
% \label{acquisition function}
% \begin{aligned}
% A^{(i)} &= \operatorname{Var}\left[L^{(i)}\right] \\
% & \approx \sum_{j=0}^{n-1} \left( \nabla_{\hat{t}_{j}} L^{(i)}\right)^2 \operatorname{Var}(t_j) + \sum_{j=0}^{n-1} \left( \nabla_{\hat{i}_{j}} L^{(i)}\right)^2 \operatorname{Var}(i_j) \\
% & \quad + 2 \sum_{j < k} \left( \nabla_{\hat{t}_{j}} L^{(i)}\right)\left( \nabla_{\hat{t}_{k}} L^{(i)}\right) \operatorname{Cov}(t_j, t_k) \\
% & \quad + 2 \sum_{j < k} \left( \nabla_{\hat{i}_{j}} L^{(i)}\right)\left( \nabla_{\hat{i}_{k}} L^{(i)}\right)  \operatorname{Cov}(i_j, i_k) \\
% & = \sum_{j=0}^{n-1} \left( \nabla_{\hat{t}_{j}} L^{(i)}\right)^2 \operatorname{Var}(t_j) + \sum_{j=0}^{n-1} \left( \nabla_{\hat{i}_{j}} L^{(i)}\right)^2 \operatorname{Var}(i_j) \\
% \end{aligned}
% \end{equation}

\begin{equation}
\label{acquisition function}
\begin{aligned}
A^{(i)} &= \operatorname{Var}\left[L^{(i)}\right] \\
& \approx \sum_{j=0}^{n-1} \left( \nabla_{\hat{t}_{j}} L^{(i)}\right)^2 \operatorname{Var}(t_j) + \sum_{j=0}^{n-1} \left( \nabla_{\hat{i}_{j}} L^{(i)}\right)^2 \operatorname{Var}(i_j) \\
\end{aligned}
\end{equation}

Here the co-variances between statistical moments are zeros. In the \(i\)-th sample, the gradient $\nabla_{\hat{t}_j} L^{(i)}$ or $\nabla_{\hat{i}_j} L^{(i)}$ indicates how sensitive the loss is to small changes in the time point $t_j$ or current point $i_j$, $j<n$. In \ref{init}, we uniformly extend all time and current data to a consistent length of n points through linear interpolation. This sensitivity is then weighted by the variance of the prediction at that time or current point, reflecting the contribution of each time or current point’s uncertainty to the overall uncertainty of the loss.

In the GPR models, our objective is to estimate the uncertainty of the entire sample based on the uncertainty of each predicted value. To achieve this, we integrate perturbation theory into variational analysis to derive the Root Mean Square Error (RMSE) between the predicted voltage waveform and the simulated waveform concerning a specific predicted value. Using \( \hat{t}_0 \) as an example, where \( \hat{t}_0 \) represents the predicted initial time point of the voltage waveform and serves as the initial reference for prediction accuracy analysis, the process involves a detailed interpolation procedure.

When calculating the gradients of \( L^{(i)} \) with respect to each predicted time point, such as \( \hat{t}_0, \hat{t}_1 \), and so forth, both the predicted waveform and the waveform at a time \( \delta \) added to the specific time point undergo an interpolation process. This interpolation ensures that both waveforms are aligned over the same \( n+1 \) time points, facilitating the calculation of the RMSE across these corresponding voltage values.

Conversely, for gradients with respect to each predicted current value, the process simplifies as the time points for the predicted waveform and the waveform after a change \( \delta \) in the current value remain consistent. Since the time points do not vary, the RMSE can be directly computed using the voltage values corresponding to these \( n \) consistent time points.

The gradient of \( L^{(i)} \) with respect to \( \hat{t}_0 \) is expressed in Eq \ref{tidu}. Since the simulated voltage waveforms of the candidate samples are not available, we employ an approximation to avoid direct computation with the unknown \( V^{(i)*} \). Similarly, the gradients of \( L \) with respect to each predicted time point, such as \( \hat{t}_0, \hat{t}_1 \), and so forth, and the corresponding currents can be calculated using the same method.

\begin{equation}
\begin{aligned}
\label{tidu}
\nabla_{\hat{t}_0} L^{(i)} &= \lim_{\delta \rightarrow 0} \frac{1}{\delta} \left[ \operatorname{RMSE}\left(\text{V}^{(i)}(\hat{t}_0+\delta) - \text{V}^{(i)*}\right) \right. \\
&\quad \left. - \operatorname{RMSE}\left(\text{V}^{(i)}(\hat{t}_0) - \text{V}^{(i)*}\right) \right] \\
&\approx \lim_{\delta \rightarrow 0} \frac{1}{\delta} \left[ \operatorname{RMSE}\left(\text{V}^{(i)}(\hat{t}_0+\delta) - \text{V}^{(i)}(\hat{t}_0)\right) \right]
\end{aligned}
\end{equation}

\begin{figure*}[!t]
    \centering
    \includegraphics[width=0.95\textwidth]{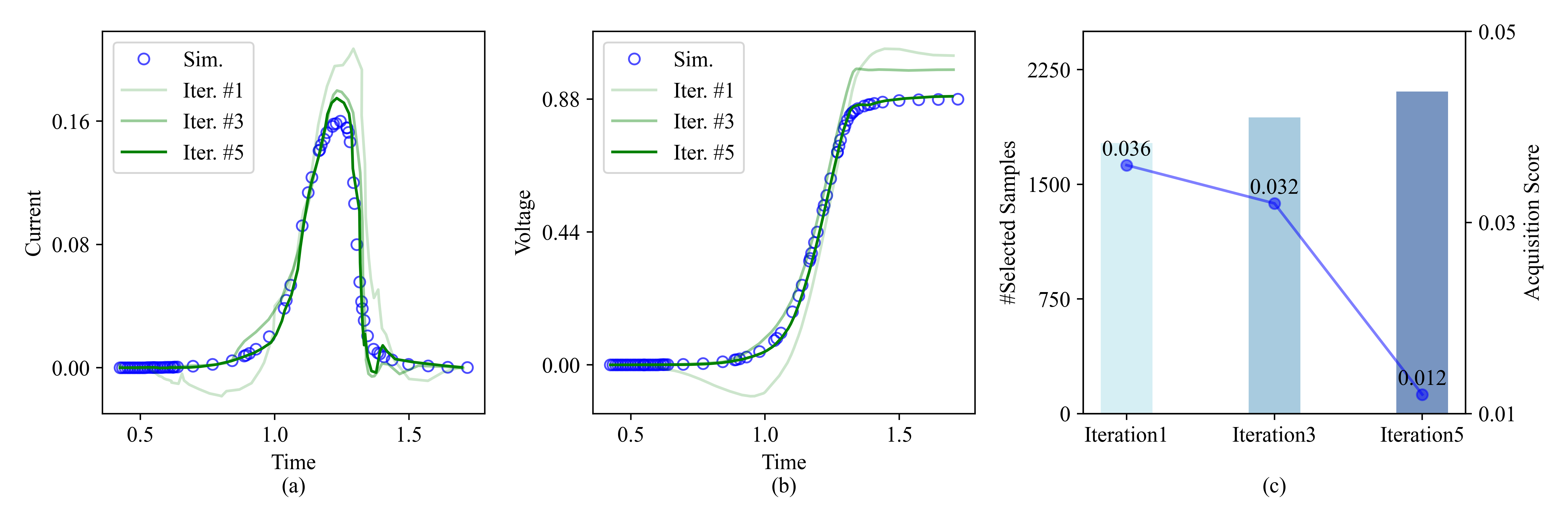} % Use the correct image file name
    \caption{Iterative Process in Waveform Estimation for INV X12 Cell at TT corner, 0.88v, 0$^{\circ}$C . (a)Current Waveform Variation. (b)Voltage Waveform Variation. (c)Acquisition Score and Selected Samples Trends.}
    \label{fig:iter_comp}
\end{figure*}

\section{EXPERIMENTS RESULTS AND DISCUSSION}
% (一页半）
\label{result}

\subsection{Experiment Setup}
\label{Setup}
The proposed cell characterization framework for CCS model is implemented in PyTorch 
%and assessed by HSPICE 
on a Linux machine with NVIDIA 3090 GPU and 8-core Intel Xeon 6348 CPU.
It was 
%, each machine running 8 tasks simultaneously and 
validated with TSMC 22nm process for 57 cells with different functions and driving strengths shown in Table I under nine PVT corners. 
In this work, the commercial tool \textit{Liberate} was utilized to generate the training data and the baseline for comparison.
%for the prediction of output current as detailed in Table 4, where the golden results were generated by commercial tool Liberate. 
The validated nine PVT corners cover the voltages between 0.8V and 0.88V and the temperatures including 0°C and 25°C.
%For other algorithms, we use pure CPU machines (as they are not designed to utilize the GPU). The comparison of delay prediction accuracy and efficiency are presented and discussed from Section \ref{accuracy} to Section \ref{runtime}.

\begin{table}[ht]
\centering
\caption{Standard Cell for Validation}
\label{tab:logic_cells}
\begin{tabular}{@{}cc@{}}
\toprule
Cell Type & Drive Strengths \\
\midrule
INV         & \makecell{X2, X3, X4, X6, X8, X9, X12,\\X15, X16, X18, X20, X21, X24} \\
AN2         & X1, X2, X4, X6, X8, X16 \\
ND2, NR2    & X0, X1, X2, X3, X4, X6, X8, X16 \\
OR2, AOI21  & X0, X1, X2, X4, X6, X8, X16 \\
XOR2, AO21  & X0, X1, X2, X4 \\
\bottomrule
\end{tabular}
\end{table}

\subsection{Verification of Acquisition Function Settings}
To assess the accuracy of the settings of the Acquisition Function, we initially applied a falling input on the timing arc of INV X1, which triggered a rising output. Fig. \ref{fig:iter_comp} demonstrates the iterative refinement of the predicted output current waveform. Fig. \ref{fig:iter_comp}\textcolor{blue}{(a)} illustrates the evolution of the current waveform throughout the iterations, showing the predicted current waveform progressively converging towards the simulated current waveform as the number of iterations increases. Fig. \ref{fig:iter_comp}\textcolor{blue}{(b)} displays the corresponding changes in the voltage waveform during the iterations, with the predicted voltage waveform gradually aligning more closely with the simulated voltage waveform, nearly matching it by the fifth iteration. Fig. \ref{fig:iter_comp}\textcolor{blue}{(c)} presents the dynamics in the number of Selected Samples and the variations in the Acquisition Score throughout the iterations. Each iteration adds only 1\% more samples to the Selected Samples, leading to only a modest increase in their total number, which facilitates a rapid decrease in the Acquisition Score. Thus, the effectiveness of the Acquisition Function is confirmed, verifying that a lower Acquisition Score corresponds to a predicted waveform that more closely resembles the waveform derived from HSPICE simulations.

\subsection{Delay Prediction Accuracy Comparison}
\label{accuracy}

To demonstrate the accuracy of our proposed model, we tested it on 57 standard cells under nine PVT corners. In the identical PVT corners, the prediction accuracy of our proposed framework for standard cells is compared with prior works \cite{9394564,8988626} as shown in Table \ref{tab:accuracy} in terms of mean absolute error (MAE) and mean absolute percentage error (MAPE). 
Our method achieves an average MAE of 2.05 ps and a average MAPE of 2.27\% across all cells and PVT corners, which represents a 6.43$\times$ and 4.49$\times$ improvement in precision over the Ridge Regression method \cite{9394564} and CSM-NN method \cite{8988626}.
These enhancements not only demonstrate our framework's superior accuracy but also underscore its robustness and reliability in handling complex variability across different conditions.

\begin{table}[h!]
\centering

% 单位写外面
\caption{Comparison of Delay Accuracy. Unit: $\textit{ps}$}
\label{tab:accuracy}
\begin{tabular}{cccc}
\hline
\makecell{Cell Type} & \makecell{Ridge \\ Regression\cite{9394564}} & \makecell{CSM-NN\\ \cite{8988626}} & \makecell{Ours} \\
\midrule
INV   & 24.64 (24.18\%)     & 15.28 (15.00\%)     & 2.92 (2.87\%)    \\ 
AN2   & 6.11 (6.42\%)       & 4.36 (4.58\%)       & 1.58 (1.66\%)    \\
ND2   & 20.45 (21.76\%)     & 12.47 (13.27\%)     & 2.55 (2.71\%)    \\ 
OR2   & 6.53 (6.41\%)       & 5.55 (5.44\%)       & 1.22 (1.19\%)    \\
NR2   & 19.00 (22.18\%)     & 13.46 (15.71\%)     & 2.64 (3.08\%)    \\ 
AO21  & 4.32 (6.37\%)       & 2.48 (3.66\%)       & 1.33 (1.97\%)    \\ 
AOI21 & 18.76 (21.74\%)     & 16.08 (18.63\%)     & 2.57 (2.98\%)    \\ 
XOR2  & 5.72 (6.21\%)       & 3.95 (4.28\%)       & 1.57 (1.70\%)    \\ \hline
\textbf{Ave.} & 13.19 (14.41\%) & 9.20 (10.07\%) & \textbf{2.05 (2.27\%)}  \\
\textbf{Norm. }& \textbf{6.43$\times$}  &\textbf{4.49$\times$  }& 1.00$\times$ \\
\hline
\end{tabular}
\end{table}

\subsection{Computational and Storage Overhead Comparison}
\label{runtime}

%The aforementioned experiments demonstrate the significantly enhanced accuracy of our model compared to other models. Moving forward, 
The computational overhead is analyzed and compared in Table \ref{tab:simulation_times} and Table \ref{tab:comparison} in terms of the number of SPICE simulation and runtime respectively, which is occupied to achieve the corresponding prediction accuracy of each method shown in Table \ref{tab:accuracy}.

The numbers of SPICE simulations in Table \ref{tab:simulation_times} include that for each type of cell of all driving strengths under all PVT corners.
It can be seen that during the training of Ridge Regression model \cite{9394564} and CSM-NN model\cite{8988626}, 80\% and 30\% samples are randomly selected for each cell respectively by taking the  number of SPICE simulations required in commercial tool as baseline.
In contrast, the number of simulations for selected samples is adaptively determined for each cell according to the uncertainty quantified by acquisition score, which are less than 30\% for all validated cells with the average ratio of 27\%.

In Table \ref{tab:comparison}, the runtime for the proposed characterization framework is compared with the  simulation time occupied by  traditional commercial tool, which consist of the simulation time for selected samples and the training time for GPR models.
It can be seen from Table \ref{tab:comparison} that the simulation and training time are respectively 26.9\% and 17.5\% in average compared to that of traditional characterization method, accounting for 44.4\% in total.
Moreover,  the required time for this work further decreases with the increase of cell complexity, which is as low as 23\% for AO21 and NR2 cell.

%various methods across all PVT corners and drive strengths for different cells. While the accuracy of the Ridge Regression\cite{9394564} and CSM-NN\cite{8988626} methods approaches that of the commercial tools, both methods achieve a reduction in the number of simulations required. Our model, however, demonstrates a significant decrease in simulation frequency, requiring only 27\% of the simulations required by the commercial tools. Although the reduction in simulations with our method is slightly less than that achieved by Ridge Regression, our model offers superior accuracy, making it a more efficient and effective choice for simulation-intensive applications.

\begin{table}[h!]
\centering
\caption{Comparison of The Number of SPICE Simulation}
\label{tab:simulation_times}
\begin{tabular}{@{}ccccc@{}}
\toprule
    
\makecell{Cell Type} & \makecell{Tool} & \makecell{Ridge \\ Regression\cite{9394564}} & \makecell{CSM-NN\\ \cite{8988626}} & \makecell{Ours} \\
\midrule
INV & 8424   & 6739  & 2527 & 2527 (30\%) \\
AN2 & 7776   & 6221  & 2333 & 2022 (26\%) \\
ND2 & 10368  & 8294  & 3110 & 2903 (28\%) \\
OR2 & 10368  & 8294  & 3110 & 2799 (27\%) \\
NR2 & 10368  & 8294  & 3110 & 2385 (23\%) \\
AO21 & 12960 & 10368 & 3888 & 2981 (23\%) \\
AOI21& 22680 & 18144 & 6804 & 6804 (30\%) \\
XOR2 & 10368 & 8294  & 3110 & 2592 (25\%) \\
\hline
\textbf{Ave.$^\ast$ } & \textbf{100\%} & \textbf{80\%} & \textbf{30\%} & \textbf{27\%}\\
\bottomrule
\end{tabular}
\caption*{$^\ast$ Percentage of the number of simulations for training by taking that for commercial tool as baseline.}
\end{table}
\begin{table}[h]
\centering
\caption{Runtime Comparison (s)}
\label{tab:comparison}
\begin{tabular}{ccccc}
\hline
\multirow{2}{*}{Cell Type} & \multirow{2}{*}{Tool} & \multicolumn{3}{c}{Ours} \\
\cline{3-5} 
& & SPICE & Train & Total \\
\hline
INV     & 144.7 & 42.1 (29.1\%) & 87.5 (60.5\%) & 129.6 (92.3\%) \\
AN2     & 210.7 & 53.8 (25.5\%) & 52.5 (24.9\%) & 106.3 (51.4\%) \\
ND2     & 246.1 & 67.4 (27.4\%) & 70.0 (28.4\%) & 137.4 (57.1\%) \\
OR2     & 233.8 & 61.7 (26.4\%) & 61.3 (26.2\%) & 123.0 (53.8\%) \\
NR2     & 248.9 & 56.1 (22.5\%) & 26.3 (10.5\%) & 82.3  (33.8\%) \\
AO21    & 257.7 & 57.8 (22.4\%) & 26.3 (10.2\%) & 84.0  (33.4\%) \\
AOI21   & 1008.2 & 299.1 (29.7\%) & 87.5 (8.7\%) & 386.6 (38.8\%) \\
XOR2    & 243.8 & 59.6 (24.5\%) & 43.8 (17.9\%) & 103.4 (43.3\%) \\
\hline
\textbf{Ave.} & \textbf{100.0\%} & \textbf{26.9\%} & \textbf{17.5\%} & \textbf{44.4\%} \\
\hline
\end{tabular}
\end{table}

%The data in Table IV highlights the efficiency of our model, which significantly reduces the average simulation time compared to traditional methods. Specifically, our model reduces the simulation time by an average of 231.2 seconds across various cell types. This reduction in simulation time is 4.1 times greater than the average training time required for our model, which is only 56.9 seconds. This demonstrates that our approach not only speeds up the simulation process significantly but also does so with a much smaller time investment for training the model. 

\begin{figure}[t]
    \centering
    \includegraphics[width=\columnwidth]{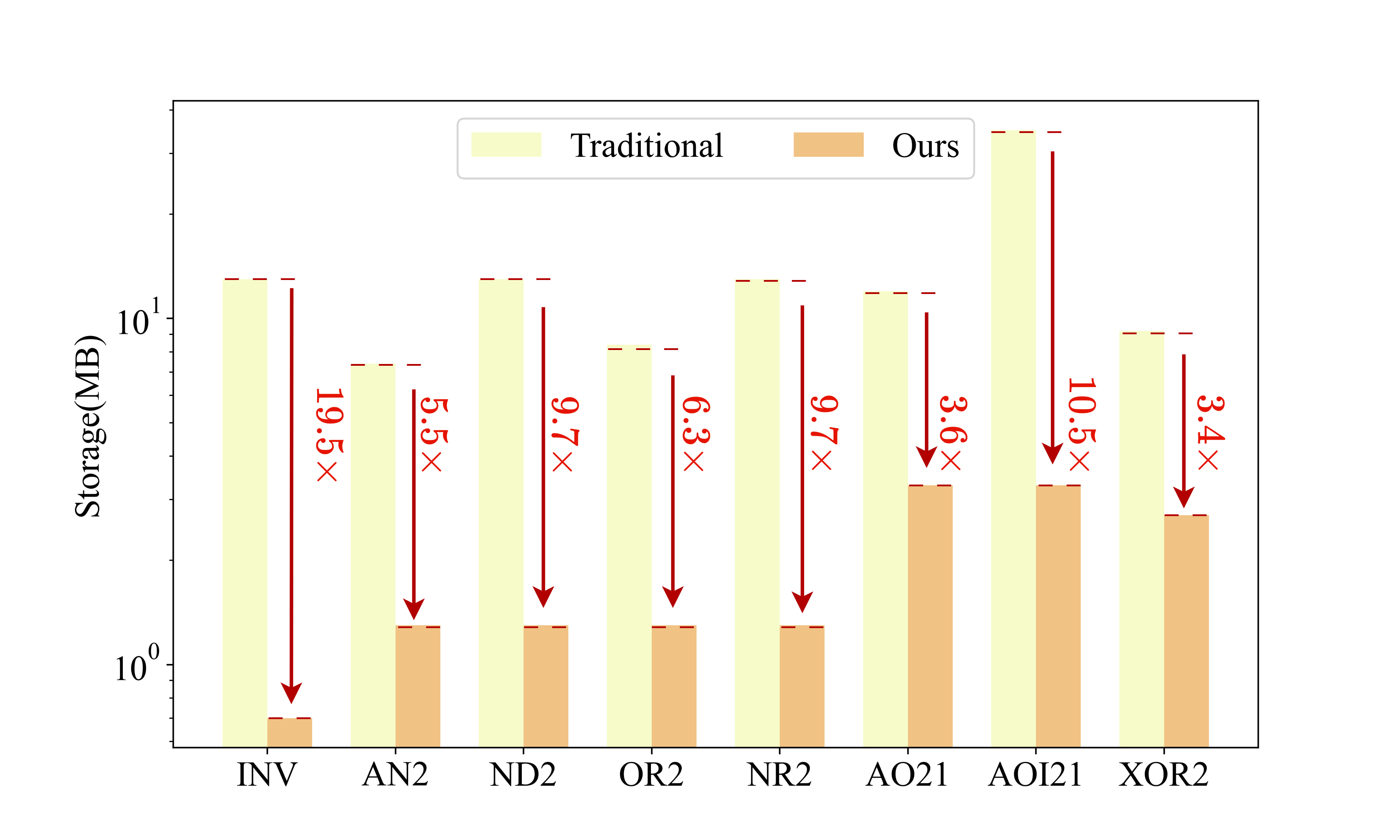} % Use the correct image file name
    \caption{Comparison of storage overhead for CCS model. }
    \label{fig:Storage}
\end{figure}

Fig. \ref{fig:Storage} demonstrates the required storage for each cell with this work and traditional library format.
In traditional, the points of current waveform for specific function of cell are stored as lookup table (LUT) for each PVT corners and driving strengths, suffering from tremendous storage overhead without the consideration of correlation among these conditions.
% With the proposed framework, the current waveforms are saved as the parameters of GPR models, from which the current and time values could be predicted by input features with trivial computational cost for inference.
With the proposed framework, the GPR models are utilized to predict current and time values based on input features. This approach enables efficient inference with trivial computational cost.
As compared in Fig. \ref{fig:Storage}, the reduction of storage rages from 3.4$\times$ to 19.5$\times$ for all validated cells, which is highly related to the number of PVT corners and driving strengths.
Since 11 driving strengths of INV cell are validated in this work, as can be seen in Table \ref{tab:logic_cells}, it achieves the maximum storage saving.

%We have extracted the current and time data for all drive strengths cells of different cell types under nine PVT corners from the standard cell libraries. It is important to note that the number of time points describing each waveform varies, highlighting the inconsistency in waveform data across different conditions. This extraction allows us to determine the storage size for each cell type and the corresponding model size, as shown in Figure 4. Our model achieves substantial reductions in storage requirements across various standard cells when compared to commercial tools. The efficiency gains range from as high as 19.5 times less storage for the INV cell, showcasing the model's maximum efficiency, to at least 3.4 times less storage for the XOR2 cell, highlighting the consistent effectiveness across different cell types. Notably, even at its least efficient, our model still provides significant storage savings, illustrating its broad applicability and potential to enhance operational efficiencies in semiconductor design.

\section{Conclusion}
\label{Conclusion}
% 三五行
In this study, we introduced GPR models with AL that efficiently and accurately characterizes standard cells using the composite current source model. Our model demonstrated superior accuracy, achieving an average absolute error of 2.05 ps and a relative error of 2.27\%. 
Compared to commercial tools, our model requires only an average of 27\% of the simulations, significantly reducing the number of SPICE simulation.
Furthermore, the model required drastically less storage—up to 19.5 times less than that required by commercial tools. This innovative approach significantly outperforms existing methods and commercial tools in terms of precision, speed, and storage costs.

\bibliographystyle{unsrt}

\bibliography{ref}

\end{document}